\documentclass[10pt,twocolumn,letterpaper]{article}

\usepackage[pagenumbers]{cvpr} 

\usepackage{multirow}
\usepackage{bm}
\usepackage[table]{xcolor}
\usepackage{xcolor,colortbl}
\usepackage{cuted}
\usepackage{makecell}
\usepackage{caption}

\makeatletter
\DeclareRobustCommand\onedot{\futurelet\@let@token\@onedot}
\def\@onedot{\ifx\@let@token.\else.\null\fi\xspace}
\def\eg{\emph{e.g}\onedot} 
\def\ie{\emph{i.e}\onedot}

\def\etal{\emph{et al}\onedot}
\renewcommand{\paragraph}{%
	\@startsection{paragraph}{4}{\z@}%
	{0.1em \@plus 0.5ex \@minus 0.2ex}{-1em}%
	{\normalsize\bf}%
}
\newcommand\rurl[1]{%
  \href{https://#1}{\nolinkurl{#1}}%
}
\definecolor{LightRed}{rgb}{0.96,0.92,0.92}

\newcommand{\by}{\bm{y}}

\newcommand{\bI}{\bm{I}}

\newcommand{\bC}{\bm{C}}
\newcommand{\bF}{\bm{F}}

\definecolor{cvprblue}{rgb}{0.21,0.49,0.74}
\usepackage[pagebackref,breaklinks,colorlinks,allcolors=cvprblue]{hyperref}
\usepackage{fontawesome}

\usepackage{float}
\usepackage[capitalize]{cleveref}
\crefname{section}{Sec.}{Secs.}
\Crefname{section}{Section}{Sections}
\Crefname{table}{Table}{Tables}
\crefname{table}{Tab.}{Tabs.}

\title{3D-Aware Multi-Task Learning with Cross-View Correlations \\for Dense Scene Understanding}

\author{
Xiaoye Wang$^{\clubsuit}$\thanks{Work done while Xiaoye was an intern with Wei-Hong.} \quad 
Chen Tang$^{\diamondsuit}$ \quad 
Xiangyu Yue$^{\diamondsuit}$ \quad 
Wei-Hong Li$^{\spadesuit}$\thanks{Corresponding Author} \\
$^{\clubsuit}$University of Cambridge\quad
$^{\diamondsuit}$The Chinese University of Hong Kong\quad 
$^{\spadesuit}$University of Bristol\\
\faGithub~\url{github.com/WeiHongLee/CrossView3DMTL}
}

\begin{document}

\maketitle

\begin{strip}
    \centering
    \begin{minipage}{0.94\linewidth}
        \centering
        \includegraphics[width=1\textwidth]{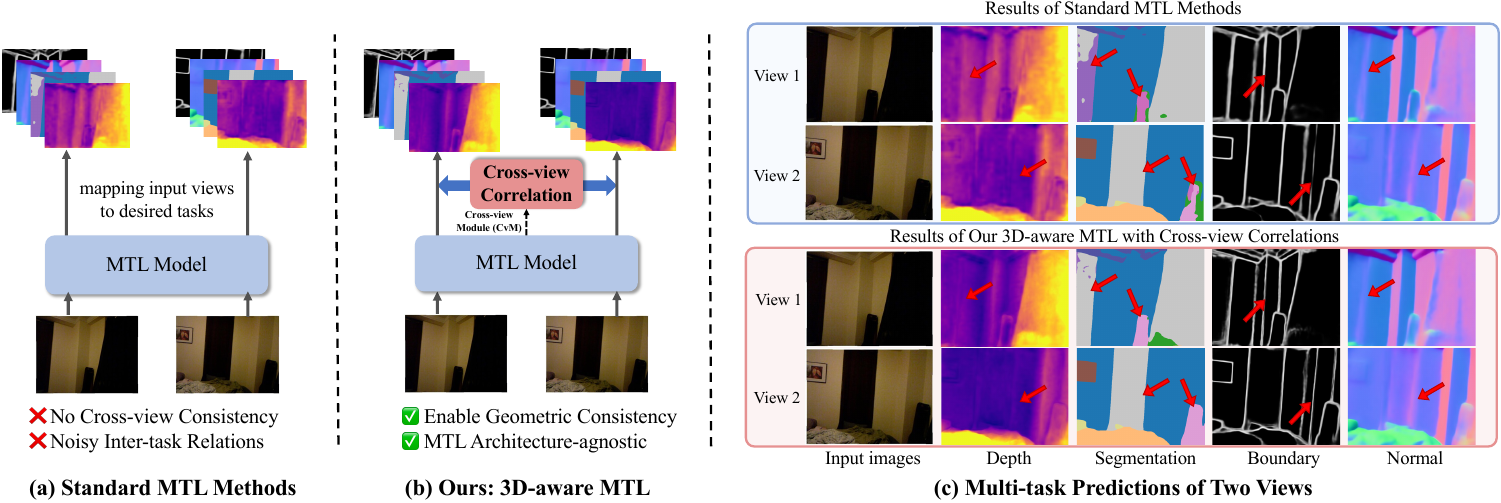}
    \end{minipage}
    \vspace{-0.3cm}
    \captionof{figure}{
    \textbf{Comparison between standard MTL and our 3D-aware MTL framework.} (a) Standard MTL relies solely on 2D per-pixel supervision, while (b) our approach incorporates geometric consistency through a lightweight cross-view module (CvM). (c) Using DINOv3~\cite{simeoni2025dinov3} as the encoder, standard MTL (top) shows cross-view ambiguities (highlighted by arrows), e.g., inconsistent curtain segmentation, leading to reduced inter-task coherence. In contrast, our method yields more consistent predictions across both views and tasks.
    }
    \vspace{-0.1cm}
    \label{fig:3dmtl}
\end{strip}

\begin{abstract}
This paper addresses the challenge of training a single network to jointly perform multiple dense prediction tasks, such as segmentation and depth estimation, i.e., multi-task learning (MTL).
Current approaches mainly capture cross-task relations in the 2D image space, often leading to unstructured features lacking 3D-awareness. We argue that 3D-awareness is vital for modeling cross-task correlations essential for comprehensive scene understanding. We propose to address this problem by integrating correlations across views, i.e., cost volume, as geometric consistency in the MTL network. Specifically, we introduce a lightweight Cross-view Module (CvM), shared across tasks, to exchange information across views and capture cross-view correlations, integrated with a feature from MTL encoder for multi-task predictions. This module is architecture-agnostic and can be applied to both single and multi-view data. Extensive results on NYUv2 and PASCAL-Context demonstrate that our method effectively injects geometric consistency into existing MTL methods to improve performance. 
\vspace{-0.43cm}
\end{abstract}

\section{Introduction}\label{sec:intro}

One central focus of multi-task learning (MTL) in computer vision~\cite{vandenhende2021multi} is to jointly solve multiple visual tasks within a single network. By sharing the majority of model parameters, MTL models effectively reduce computational cost and memory capacity while exploiting cross-task inductive biases \cite{caruana1997multitask, vandenhende2021multi, yu2024unleashing, li2023multi}. 
This multi‑task collaborative framework resonates deeply with a variety of real‑world applications \cite{yu2024unleashing}, such as robotic automation, where depth estimation and spatial awareness for obstacle avoidance are combined with semantic segmentation and other scene understanding tasks for object localization~\cite{ainetter2021depth}.

However, building an MTL model that performs consistently well for all desired tasks remains a challenging problem as it requires the MTL model to maintain a good balance between shared and task-specific features~\cite{li2023multi}. Modern deep learning-based MTL methods have explore various architectures innovations to address this: Liu \etal~\cite{liu2019end} introduce task-specific attention modules with a fully shared feature encoder for more flexible feature sharing; Vandenhende \etal \cite{vandenhende2021multi} and Bruggemann \etal \cite{vandenhende2021multi} design cross-task attention modules to capture inter-task relations from multi-scale features; Recent transformer-based methods have advanced MTL through techniques such as high-resolution multi-scale decoding~\cite{ye2022inverted}, prompt learning~\cite{ye2022taskprompter,wang2024tsp}, task-specific experts~\cite{ye2023taskexpert,yang2024multi,chen2023mod,fan2022m3vit} and multi-teacher knowledge distillation~\cite{ranzinger2024radio,heinrich2025radiov2,lu2024swiss}.

Despite advancements, most existing MTL methods predominantly rely on mapping 2D images to high-dimensional features and per-pixel supervision (as shown in ~\cref{fig:3dmtl} (a)), resulting in unstructured features. This unstructured feature space, coupled with a lack of explicit mechanisms for modeling spatial consistency, can lead to noisy inter-task relations and degraded performance~\cite{li2023multi,zheng2023multi} (\cref{fig:3dmtl} (c) top). In response, pioneering work like 3DMTL~\cite{li2023multi} explores integrating 3D regularization into MTL by proposing a structured 3D-aware regularizer that projects shared features into a 3D feature space and decodes multiple tasks via differentiable rendering. Another approach, MuvieNeRF~\cite{zheng2023multi} recasts multi-task dense prediction as multi-view synthesis, embedding both cross-view and cross-task attention within a NeRF~\cite{mildenhall2021nerf} backbone to synthesize multiple scene properties. However, despite utilizing multi-view data for 3D-awareness via feature projection and rendering, 3DMTL~\cite{li2023multi} does not directly extract and integrate multi-view geometric cues into its shared representations. Meanwhile, MuvieNeRF~\cite{zheng2023multi} requires multi-view data and camera parameters during inference, which limits its scalability for real-world MTL applications.

On the other hand, recent work on multi-view scene reconstruction, such as VGGT~\cite{wang2025vggt}, MVSplat~\cite{chen2024mvsplat}, and DepthSplat~\cite{xu2025depthsplat}, demonstrates the success of leveraging multiple views for building robust 3D representations. MVSplat~\cite{chen2024mvsplat} effectively reconstructs high-fidelity 3D scenes by efficiently processing sparse multi-view images with 3D Gaussian Splatting. Complementing this, DepthSplat~\cite{xu2025depthsplat} integrates depth information from a pretrained depth estimation model with Gaussian Splatting to achieve superior 3D reconstruction quality. VGGT~\cite{wang2025vggt} leverages a unified Transformer architecture enhanced with visual geometry grounding to improve 3D understanding from multi-view inputs and multiple downstream 3D tasks. 
However, these methods primarily focus on scene reconstruction or 3D representation and are not directly designed to enhance multi-task learning that jointly performs multiple dense prediction tasks from single view image input, including depth estimation, boundary detection, surface normal estimation, and semantic segmentation, making their integration into MTL unclear. 
This leads us to a critical question for current MTL frameworks:
\begin{center}
\vspace{-0.05cm}
    \textit{
        Can we introduce cross-view correlations into MTL to ensure high geometric consistency across vision tasks?
    }
\end{center}

Motivated by this insight, we propose a 3D-aware multi-task learning framework that tightly integrates the design principles of MTL and 3D reconstruction, as shown in ~\cref{fig:3dmtl} (b). 
Our approach augments monocular multi-task learning feature encoder with multi-view geometric cues via a dedicated geometric pathway, which we term the Cross-view Module (CvM), allowing the model to learn representations that are simultaneously task-aligned and geometry-aware (\cref{fig:3dmtl} (c)). The CvM consists of three sequential components: \emph{(i)} a spatial-aware encoder for extracting geometric and spatial primitives, \emph{(ii)} a multi-view transformer that ingests these encoded features for relational interactions, and \emph{(iii)} a cost volume module that reconstructs cross-view correlations as a geometric representation. Crucially, the spatial-aware encoder operates independently of the main monocular MTL pathway, allowing it to leverage strong inductive biases for spatial locality to explicitly capture geometry-rich cues. Its features are then passed into the multi-view transformer, where intra- and cross-view attention forge robust geometry-aware representations, culminating in the construction of a cost volume that establishes dense correspondence across views. These 3D-aware features complement the original monocular MTL features and are concatenated with them before being passed into lightweight, task-specific decoder heads. Our method supports both MTL training on multi-view data (or video inputs) and single-view, while only a single image is needed for inference, making it broadly applicable in practice.

To summarize, our main contributions are as follows:
\begin{itemize}
    \item Unlike prior work that primarily focuses on learning direct mappings between input images and desired task ground-truths, we propose to enable 3D-aware MTL by integrating cross-view correlation, \ie, cost volume, as a geometric consistency into multi-task learning through an introduced multi-view module.
    \item Our method is architecture-agnostic, allowing seamless plugged into various existing MTL architectures to enhance their performance.
    \item Our approach is applicable to both single and multi-view data during training, yet requires only a single image for inference, eliminating the need for camera parameters at deployment.
    \item Extensive experimental results demonstrate the effectiveness and superiority of our proposed 3D-aware multi-task learning framework on standard NYUv2 and PASCAL-Context benchmarks.
\end{itemize}

\section{Related Work}\label{sec:rel}
\subsection{Multi-task Learning}

Multi-task Learning (MTL)~\cite{caruana1997multitask} aims at learning a single network that jointly estimates accurate predictions for multiple desired tasks. Recent research of multi-task learning in computer vision can be broadly divided into two categories~\cite{vandenhende2021multi,ruder2017overview}. 
The first group aims at addressing the unbalanced optimization issues - each task's loss function often exhibits distinct scales and convergence behaviors, jointly minimizing them can lead to optimization conflicts and performance degradation. To address this issue, prior work proposes to estimate loss weights dynamically~\citep{kendall2018multi,liu2019end,guo2018dynamic,chen2018gradnorm,lin2019pareto,sener2018multi,liu2021towards}, mitigating conflicts between gradient conflicts~\cite{yu2020gradient,liu2021conflict,chen2020just,chennupati2019multinet++,suteu2019regularizing} or aligning features with single-task models~\citep{li2020knowledge,li2022universal}.

Our work is more related to the second one which aims at designing architectures~\citep{kokkinos2017ubernet,ruder2019latent,vandenhende2019branched,liang2018evolutionary,bragman2019stochastic,strezoski2019many,xu2018pad,zhang2019pattern,bruggemann2021exploring,bilen2016integrated,zhang2018joint,xu2018pad} that better share information across tasks 
by cross-task attention mechanisms~\cite{misra2016cross}, task-specific attention modules~\citep{liu2019end,bhattacharjee2023vision}, cross-tasks feature interaction~\citep{ye2022inverted,vandenhende2020mti}, gating strategies or {mixture of experts modules~\citep{bruggemann2020automated,guo2020learning,chen2023mod,fan2022m3vit,ye2023taskexpert}, visual prompting~\citep{ye2022taskprompter,liu2023hierarchical}} and distillation of multiple visual foundation models~\cite{ranzinger2024radio,heinrich2025radiov2,lu2024swiss}. However, these methods mainly capture cross-task relations within the 2D space, and two recent works~\cite {li2023multi,zheng2023multi} show that 3D-awareness is vital for learning more structured features that are shared and beneficial for all tasks by using NeRF as a decoder or a 3D-aware regularizer. However, MuvieNeRF~\cite{zheng2023multi} requires multiple views and ground-truth camera parameters, limiting the usages for practical scenarios. While 3DMTL~\cite{li2023multi} does not require camera parameters during inference and can be applied to standard MTL settings with single-view input, it is shown that the method has limited capacity of leveraging multi-view data for enhancing the 3D-awareness. 

Unlike existing methods, we propose to enable MTL to be 3D-aware by equipping the standard MTL encoder with a cross-view module capturing cross-view relations, allowing the model to learn representations that are simultaneously task-aligned and geometry-aware. Additionally, our method is architecture agnostic and can be applied to both single and multi-view data without requiring camera parameters during inference.

\subsection{3D Scene Reconstruction and Synthesis}
Our approach is also related to methods that learn 3D scene representations for multi-view scene reconstruction and synthesis~\citep{lombardi2019neural,mildenhall2021nerf,chan2021pi,cai2022pix2nerf,gu2021stylenerf,wang2024dust3r,wang2025vggt,chen2024mvsplat,leroy2024grounding,yang2025fast3r,zhang2025flare}.
Earlier works~\citep{mildenhall2021nerf,kerbl20233d} in this field represent only a single scene per model, require many calibrated views, or are not able to perform other tasks than novel view synthesis such as semantic segmentation, depth estimation. Zhi \etal~\cite{zhi2021place} extend the standard NeRF pipeline through a parallel semantic segmentation branch to jointly encode semantic information of the 3D scene, and obtain 2D segmentations by rendering the scene for a given view. Panoptic Neural Fields~\citep{kundu2022panoptic} predict a radiance field encoding color, density, instance, and category labels for any 3D point by combining multiple encoders for both background and individual object instances. However, this approach is limited to predicting these tasks on novel views of previously seen scenes. Consequently, it cannot be applied to entirely new scenes without additional training and is constrained to handling only rigid objects.

PixelNeRF~\citep{yu2021pixelnerf} and PixelSplat~\cite{charatan2024pixelsplat} condition a NeRF~\citep{mildenhall2021nerf} or Gaussian Splatting~\cite{kerbl20233d} on image inputs through an encoder, allowing for the modeling of multiple scenes jointly and generalizes to unseen scenes, however, the work focuses only on synthesizing novel views. MVSplat~\cite{chen2024mvsplat} further improves PixelSplat~\cite{charatan2024pixelsplat} by efficiently incorporating cross-view correlations to improve the scene reconstruction. Building on this, DepthSplat~\cite{xu2025depthsplat} proposes to leverage depth prediction from a single-view depth model for further improving the quality of 3D scene reconstruction. More recently, Wang \etal~\cite{wang2025vggt} propose VGGT that directly infers a set of 3D scene attributes from multiple views using a single, efficient feed-forward transformer. 

In contrast to these methods that focus on scene reconstruction or synthesis, our work focuses on joint learning of dense vision problems in novel scenes and leverages cross-view correlation as a geometry cue to bring a beneficial structure to the learned representations.
Our method can be trained from single-view or multi-view inputs and is not limited to a fixed architecture or specific set of tasks.

\section{Methodology}\label{sec:method}
\begin{figure*}[t!]
    \centering
    \includegraphics[width=0.92\textwidth]{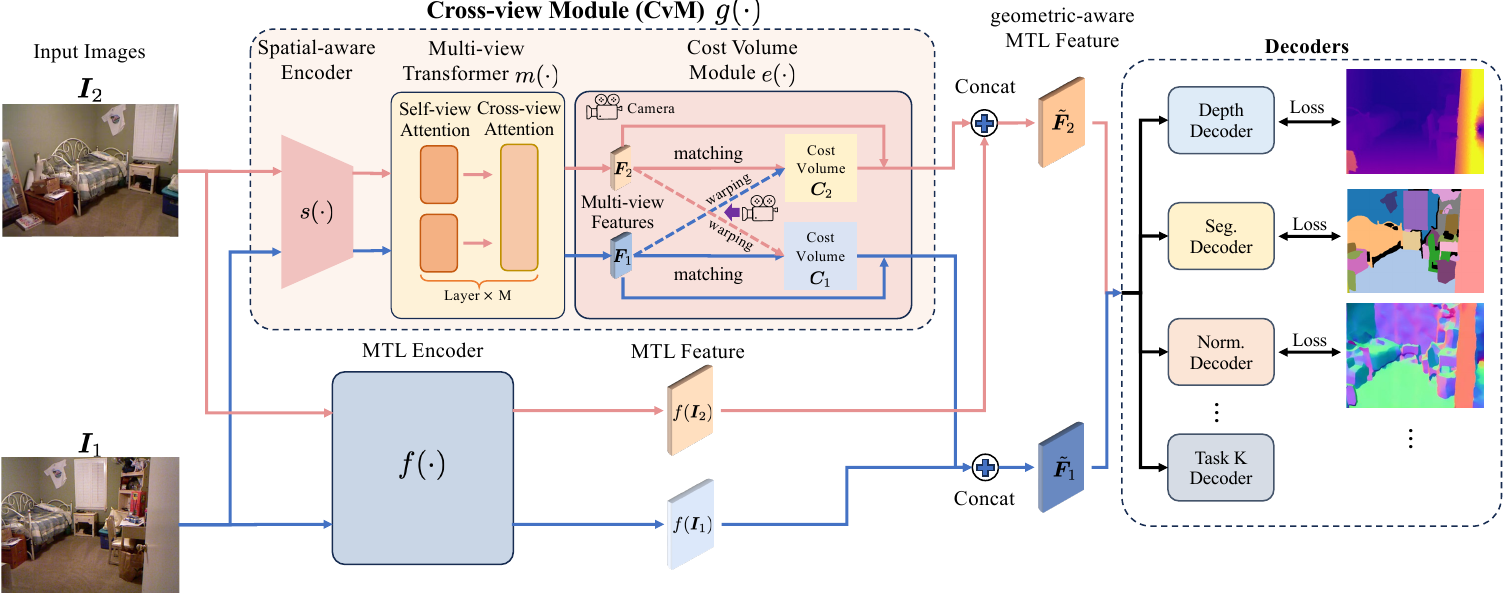}
    \vspace{-0.3cm}
    \caption{\textbf{Illustration of our method for integrating cross-view correlation for enabling 3D-aware MTL.} 
    Given an image $\bI_1$, we feed it and its neighbour view $\bI_2$ into the MTL encoder $f(\cdot)$ and extract the MTL features $f(\bI_1)$ and $f(\bI_2)$. In parallel, our lightweight cross-view module (CvM) $g(\cdot)$ takes as input both views. In CvM, a spatial-aware encoder $s(\cdot)$ encodes geometric-biased features, followed by a multi-view transformer $m(\cdot)$ that enables information exchanged across views and outputs cross-view features $\bF_1$ and $\bF_2$. A cost volume module $e(\cdot)$ then converts $\bF_1$ and $\bF_2$ to the cost volume $\bC_1$ and $\bC_2$ by warping the feature from one view to another given their relative camera parameters and matching features across views. Finally, both cost volume and cross-view feature are concatenated with the MTL features, forming the geometric-aware MTL feature $\Tilde{\bF}_1$ and $\Tilde{\bF}_2$ for estimating predictions of multiple dense vision tasks.
    }
    \vspace{-0.1cm}
    \label{fig:pipeline}
\end{figure*}

\subsection{Multi-task Learning}

In multi-task learning (MTL), we aim to train a single model that takes an RGB image as input and simultaneously predicts multiple dense output targets, such as depth, edges, semantic labels, and surface normals. Formally, with an input image $\bI \in \mathbb{R}^{H \times W \times 3}$, the goal is to estimate a set of task-specific outputs $\mathcal{Y} = \{\by_1, \dots, \by_T\}$ corresponding to $T$ different tasks.

A common approach to MTL is to employ a shared encoder with $T$ task-specific decoders architecture, where a feature extractor $f(\cdot)$ maps the input image to a latent representation $f(\bI) \in \mathbb{R}^{H' \times W' \times C}$. This shared representation is then processed by $T$ lightweight task-specific decoders $\{h_t\}_{t=1}^T$ to generate the task predictions $\hat{\by}_t = h_t \circ f(\bI)$. Such designs exploit the redundancy between related tasks and improve training efficiency by sharing features across tasks.

The model is typically trained on a single-view labeled dataset $\mathcal{D}$ with $N$ image-label pairs by jointly minimizing multiple losses:
\begin{equation}\label{eq:mtl}
\min_{f, \{h_t\}_{t=1}^T} \frac{1}{N} \sum_{(\bI, \mathcal{Y}) \in \mathcal{D}} \: \sum_{\by_t \in \mathcal{Y}} \ell_t(h_t \circ f(\bI)), \by_t),
\end{equation}
where $\ell_t$ denotes the task-specific loss function, e.g., cross-entropy for segmentation, $L_1$ loss for depth estimation.

\subsection{MTL with Cross-view Correlations}
While existing multi-task learning benefits from feature sharing, it relies on single-view 2D images, and \emph{its inherent lack of 3D awareness often results in geometric inconsistencies between related tasks}. To mitigate this, one straightforward approach is to include multi-view data (e.g., video sequences) for training multi-task learning (MTL) models to improve geometric consistency. However, simply mapping multi-view inputs to the desired task ground-truths does not ensure consistency across views of the same scene, which is a crucial geometric cue for scene understanding.

To address this, we propose a 3D-aware multi-task learning framework that augments the shared encoder $f(\cdot)$ with a lightweight cross-view module (CvM) $g(\cdot)$—\emph{shared across all tasks}—comprising: (i) a spatial-aware encoder that extracts geometry-biased features from paired views; (ii) a multi-view transformer that performs self/cross-attention to exchange information and produce cross-view enhanced features; and (iii) a differentiable cost volume builder that warps and matches features across depth hypotheses to explicitly encode cross-view correlations. The resulting geometric representation is fused with features from $f(\cdot)$ to obtain task-specific predictions (\cref{fig:pipeline}), enforcing cross-view geometric consistency and improving 3D coherence across tasks. 
Fig.~\ref{fig:pipeline} illustrates the overall pipeline of our method. 
The detailed design of our CvM module is as follows:

\paragraph{Spatial-aware encoder} provides geometry-biased features from each view that decoupled from monocular MTL cues, to serve as clean inputs for cross-view correlation modeling shared by all tasks. 
More specifically, given an image $\bI_1$, we feed it and its neighboring views $\{\bI_i\}_{i=2}^V$ into the MTL encoder to extract the MTL features $\{f({\bI_i})\}_{i=1}^V$. In this work, we use 1 neighbour view, \ie, $V=2$, but the method supports more views. 
One could simply utilize the MTL features for cross-view matching and regularizing cross-view consistency. However, we argue that this can lead to interference between monocular MTL and cross-view matching, leading to higher difficulty in training (shown in \cref{tab:nyuv23Ddesign}) and extending to other MTL models. 

To this end, we instead design a cross-view module that operates in parallel with the MTL encoder. This cross-view module consists of a spatial-aware encoder $s(\cdot)$ that extracts geometric-aware features, followed by a multi-view transformer $m(\cdot)$ which aggregates intra- and cross-view correspondences, and a cost volume module $e(\cdot)$ for constructing cross-view correlations as geometric representations. The spatial-aware encoder $s(\cdot)$ is implemented as a shallow ResNet-like~\cite{he2016deep} Convolutional Neural Network (CNN), similar to \cite{chen2024mvsplat,xu2023unifying}, to extract spatial-aware downsampled features of the all views $\{s(\bI_i)\}_{i=1}^{V}$.

\paragraph{Multi-view transformer} aggregates complementary intra- and cross-view context to strengthen correspondences and disambiguate occlusions/textureless regions, yielding cross-view enhanced features for subsequent geometry construction. 
Instead of directly matching spatial-aware features $\{s(\bI_i)\}_{i=1}^{V}$, we adopt a multi-view transformer, implemented as a multi-view Swin Transformer \cite{liu2021swin, xu2022gmflow, xu2023unifying, xu2025depthsplat}, consisting of stacked self- and cross-attention layers to facilitate information exchange across views. 
Within this transformer, for each view, we compute attention with respect to its neighboring views\footnote{Our method supports $V>2$ though we use $V=2$ by default. For each reference view that has more than 2 neighbour views (\ie, $V>3$), we perform cross-attention between the reference view and its top-2 nearest neighboring views, which are selected based on their camera distances to the reference view to ensure better trade-off between performance and computational efficiency.}, enabling the transformer to aggregate complementary visual cues and disambiguate challenging regions such as occlusions and textureless surfaces. To balance computational efficiency and geometric consistency, we follow a local attention design similar to Swin Transformer~\cite{liu2021swin}, where attention is restricted within spatial windows but is repeated across all scales and views. This ensures that the resulting features are both geometry-aware and scalable to large input resolutions, which is important for dense vision tasks. The output of the multi-view transformer is a set of cross-view enhanced feature maps $\{\bF_i\}_{i=1}^V = m(\{s(\bI_i)\}_{i=1}^{V})$, which are subsequently used for constructing the cost volume.

\paragraph{Cost volume module} converts learned correspondences into an explicit, differentiable 3D representation by building a depth-parameterized cost volume that enforces geometric consistency. 
Given cross-view enhanced feature maps $\{\bF_i\}_{i=1}^V$, we aim to encode the feature matching information across views as a geometric cue, shared across all dense vision tasks, to improve their performance. Following prior work in multi-view stereo~\cite{yao2018mvsnet,xu2023unifying},
we construct a differentiable cost volume to explicitly model the geometric consistency across views. To achieve this, we first define a set of $L$ candidate depth planes $\{d_1, d_2, \ldots, d_L\}$ sampled uniformly in inverse depth space. For each candidate depth $d$, the feature of one neighboring view $\bI_j$ is warped to the reference view $\bI_i$ using their camera intrinsics and relative pose, producing $\hat{\bF}_{j \rightarrow i}^{(d)}$.

For each view $\bI_i$, we then match its feature and each neighbour view at each pixel location using the dot-product similarity between the feature and the warped features for each depth candidate and aggregate over all neighbour views and all depth candidates:
\begin{equation}
\bC_i = e(\{\bF_i\}_{i=1}^V)= \frac{1}{V-1}\sum_{\substack{j=1 \\ j \neq i}}^V \sum_{d=1}^{L}\frac{\bF_i \cdot \hat{\bF}_{j \rightarrow i}^{(d)}}{\sqrt{K}},
\end{equation}
where $K$ is the channel dimension for normalization. And this yields a 3D cost volume $\bC_i \in \mathbb{R}^{H \times W \times D}$ for each view shared across all tasks.

\paragraph{Training objective.}
Finally, we concatenate the cost volume $\bC_i$ and the cross-view enhanced feature $\bF_i$ with the MTL feature $f(\bI_i)$ to form the 3D-aware multi-task feature $\Tilde{\bF}_{\bI_i}=\text{concat} \left(f \left(\bI_i \right), \bC_i, \bF_i \right)$ for estimating multi-task predictions.
We then measure the mismatch between ground-truth labels and the predictions obtained from the spatial-aware MTL feature, and jointly optimize the model by minimizing all task losses as in \cref{eq:mtl}:
\begin{equation}\label{eq:3dmtl}
\min_{f, \{h_t\}_{t=1}^T, g} \frac{1}{NV} \sum_{\{(\bI_i, \mathcal{Y}_i)\}_{i=1}^{V} \in \mathcal{D}} \: \sum_{\by_t \in \mathcal{Y}_i} \ell_t(h_t \circ \Tilde{\bF}_{\bI_i}, \by_t),
\end{equation} where $g=e \circ m \circ s$
is the cross-view module that consists the spatial-aware encoder $s$, multi-view transformer $m$ and the cost volume module $e$.

\paragraph{Training and inference with single-view inputs.}
Although our cross-view module requires at least two views as input, during inference, often only a single-view image is available. We address this by duplicating the single-view image to serve as the neighboring view, enabling multi-task predictions. We employ the same strategy for training our method on single-view datasets and found that it still performs effectively (as shown in \cref{tab:nyuv2sota} and \cref{tab:pascalsota}). We hypothesize that training the cross-view module on duplicated single-view images can prevent it from capturing spurious correlations between identical views, thereby enhancing its robustness. However, further improvements could be achieved through augmentation or multi-view image generation techniques, and we leave this for future work.

\section{Experiments}\label{sec:exp}
In this section, we first detail the benchmarks and our implementation, followed by a quantitative and qualitative analysis of our method. Please refer to the supplementary materials for more results and details.

\subsection{Datasets}

\paragraph{NYUv2\rm{~\cite{silberman2012indoor}}}
is a popular MTL benchmark consisting of 1449 indoor RGB-D images captured with a Microsoft Kinect sensor. We follow the standard split~\cite{Eigen2015Predicting} and we use 795 and 654 images for training and testing. Following the prior work~\cite{lu2024swiss,ye2022inverted}, we perform four tasks, including 40-class semantic segmentation, depth estimation, surface normal prediction, and boundary detection.

\paragraph{NYUv2 Video Frames.} Following prior work~\cite{li2023multi}, we leverage raw RGB-D video sequences from the NYUv2 dataset~\cite{silberman2012indoor}, extracting additional video frames to construct multi-view inputs. 
We also follow 3DMTL~\cite{li2023multi} and use COLMAP~\cite{schoenberger2016sfm} to estimate relative camera poses between adjacent frames. These video frames only have depth annotations and they are used for training in multi-view setting.

\paragraph{PASCAL-Context\rm{~\cite{chen2014detect}}} provides dense annotations for various visual tasks, including semantic segmentation, boundary detection, and human part segmentation. We follow~\cite{vandenhende2021multi} and also perform saliency detection, and surface normal prediction with annotation from Vandenhende \etal~\cite{vandenhende2021multi}. We adopt the standard splits~\cite{vandenhende2021multi, ye2022inverted}: 4998 images for training and 5105 images for testing. 

\subsection{Implementation Details}
Our method is architecture agnostic and can be applied to different state-of-the-art MTL methods. We apply our method to the recent SAK~\cite{lu2024swiss}, RADIO~\cite{ranzinger2024radio} from Lu \etal~\cite{lu2024swiss} and DINOv3~\cite{simeoni2025dinov3}, by attaching the cross-view module (CvM) to their encoder. 
For all experiments, we follow the identical training and evaluation protocol of prior work~\cite{lu2024swiss}.
We implement our model in PyTorch \cite{paszke2019pytorch} 
and use the same loss functions and loss weights as in~\cite{lu2024swiss,ye2022inverted,vandenhende2021multi}.
Across all experiments, we use ViT-L as the backbone for MTL encoder.
For CvM, our spatial-aware encoder produces features with 128 dimensional at $1/8$ input resolution, followed by the multi-view transformer with 6 self and cross-view attention layers. The number of depth candidates $L$ is set to be 128 to uniformly sample depth candidates from 0.0001 to 10. Please refer to supplementary for more details.

\paragraph{Evaluation Metrics.}
We follow the previous methods \cite{ye2022inverted, lu2024swiss}, measuring semantic segmentation and human parsing by the mean Intersection over Union (mIoU), saliency detection by maximum F-measure (maxF), surface normal estimation by mean error (mErr) of angles, depth estimation by Root Mean Square
Error (RMSE), and boundary detection by optimal-dataset scale F-measure (odsF) \cite{martin2004learning,pont2015supervised}.
We also report the multi-task learning performance $\Delta$MTL, \ie, average performance gains across all tasks w.r.t. to a baseline, \eg, single task learning method, as in prior work~\cite{vandenhende2021multi}.
{
\setlength{\tabcolsep}{3.2pt} 
\begin{table}[t]
	\centering
	
    \resizebox{0.48\textwidth}{!}
    {
		\begin{tabular}{lccccccccc}

		    \toprule
		     Method 
             & \makecell{Seg.\\(mIoU) $\uparrow$} 
             & \makecell{Depth\\(RMSE) $\downarrow$} 
             & \makecell{Normal\\(mErr) $\downarrow$} 
             & \makecell{Boundary\\(odsF) $\uparrow$}  
             & \makecell{$\Delta$MTL $\uparrow$} \\
            \midrule
            SAK~\citep{lu2024swiss} \emph{w/o video} & {\bf 63.18} & 0.4313 & 16.25 & 79.43 & 0.00 \\ 
            SAK~\citep{lu2024swiss} & 62.60 & 0.4093 & 16.19 & 79.58 & 1.19 \\ 
            \rowcolor{LightRed} {\bf Ours} & 62.78 & {\bf 0.4034} & {\bf 16.10} & {\bf 80.52} & {\bf 2.03}\\ 
            \midrule
            DINOv3~\citep{simeoni2025dinov3}~ \emph{w/o video} & 63.68 & 0.4113 & 15.53 & 80.10 & 0.00 \\
            DINOv3~\citep{simeoni2025dinov3} & 64.03 & 0.3954 & 15.35 & 80.52 & 1.52 \\
            3DMTL$^{*}$ & 64.25 & 0.3952 & {\bf 15.24} & 80.15 & 1.68 \\
            \rowcolor{LightRed} {\bf Ours} & {\bf 65.27} & {\bf 0.3836} & 15.35 & {\bf 81.69} & {\bf 3.09}\\ 
			\bottomrule
		\end{tabular}%
			}
		\vspace{-0.25cm}
		\caption{Quantitative comparison of our method on NYUv2 dataset + extra video frames with multiple views. $^{*}$: We reproduce 3DMTL~\cite{li2023multi} with Dinov3 backbone. $\Delta$MTL is computed using ``SAK~\cite{lu2024swiss} \emph{w/o video}'' and ``DINOv3~\cite{simeoni2025dinov3} \emph{w/o video}'' as baseline, respectively.}
		\label{tab:nyuv2mv}
\end{table}%
}
\subsection{Results}

\paragraph{MTL with Multiple Views.} 
We perform experiments on multi-view data by training models on both single-view training images and video frames on NYUv2 and evaluating models on the single-view testing set of NYUv2. 
We incorporate our method with SAK~\cite{lu2024swiss} and DINOv3~\cite{simeoni2025dinov3} in this setting, and the results are depicted in \cref{tab:nyuv2mv}.
We observe that for both SAK \cite{lu2024swiss} and DINOv3 \cite{simeoni2025dinov3}, simply introducing multi-view data for training improves MTL performance, such as depth and surface normal estimation. However, this does not fully exploit the cross-view correlations.
In contrast, by explicitly modeling spatial correspondence and aggregating cross-view features, our CvM enables more effective use of multi-view signals. 
Notably, DINOv3 with CvM trained on multi-view data achieves +1.57 over DINOv3 trained with multi-view data, and +3.09 over the DINOv3 baseline trained with single-view data.

We further compare our approach with 3DMTL~\cite{li2023multi}, which injects 3D awareness into MTL through a neural rendering regularization. 
Our CvM achieves consistently better results, improving segmentation by +1.0, boundary detection by +1.5, and reducing depth RMSE from 0.3952 to 0.3836, while achieving comparable performance on surface normal. These results strongly indicate that the cross-view correlations learned by our method effectively enhance 3D awareness and hence improve MTL framework.

{
\setlength{\tabcolsep}{3.2pt} 
\begin{table}[t]
    \centering
	
    \resizebox{0.48\textwidth}{!}
    {
		\begin{tabular}{lcccccccccc}

		    \toprule
		     \multirow{2}{*}{Method} &  Seg. & Depth & Normal & Boundary & \multirow{2}{*}{$\Delta$MTL $\uparrow$}  \\
             & (mIoU) $\uparrow$ & (RMSE) $\downarrow$ & (mErr) $\downarrow$ & (odsF) $\uparrow$ & \\
		    \midrule
            STL  & 54.19 & 0.5560 & 19.22 & 78.09 & 0.00 \\
            MTL &  52.42 & 0.5413 & 19.29 & 76.50 & -0.76 \\
            \midrule
            TaskExperts~\cite{ye2023taskexpert}  & 55.35 & 0.5157 & 18.54 & 78.40 & 3.33 \\
            BFCI~\cite{zhang2023rethinking} &  55.51 & 0.4930 & 18.47 & 78.22 & 4.46 \\
            TSP~\cite{wang2024tsp}  & 55.39 & 0.4961 & 18.44 & 77.50 & 4.07 \\
            MLoRE~\cite{yang2024multi}  & 55.96 & 0.5076 & 18.33 & 78.43 & 4.26 \\
            InvPT~\citep{ye2022inverted}  & 53.56 & 0.5183 & 19.04 & 78.10 & 1.64 \\
           3DMTL~\citep{li2023multi}  & 54.87 & 0.5006 & 18.55 & 78.30 & 3.74\\
            \midrule RADIO~\citep{ranzinger2024radio}  & 59.32 & 0.4698 & 17.46 & 79.41 & 8.95 \\
            \rowcolor{LightRed}{\bf Ours}  & {\bf 60.26} & {\bf 0.4619} & {\bf 17.34} & {\bf 80.36} & {\bf 10.20} \\
            \midrule
            SAK~\citep{lu2024swiss}  & {\bf 63.18} & 0.4313 & 16.25 & 79.43 & 14.05 \\
            \rowcolor{LightRed} {\bf Ours} & 63.12 & {\bf 0.4044} & {\bf 16.22} & {\bf 80.56} & {\bf 15.63} \\
            \midrule
            DINOv3~\citep{simeoni2025dinov3} & 63.68 & 0.4113 & 15.53 & 80.10 & 16.33 \\
            \rowcolor{LightRed} {\bf Ours} & {\bf 64.98} & {\bf 0.3909} & {\bf 15.27} & {\bf 81.58} & {\bf 18.66} \\ 
			\bottomrule
		\end{tabular}%
			}
		\vspace{-0.25cm}
		\caption{Quantitative comparison of our method to the SotA methods on NYUv2 dataset. $\Delta$MTL is computed using single-task learning ``STL'' as baseline.}
		\label{tab:nyuv2sota}
\end{table}%
}

\paragraph{Comparisons with SotAs.} 

Our method is not limited to training on multi‑view input and can be applied in a single–view setting for comparison with current state‑of‑the‑art (SotA) MTL methods. 
We compare our method with SotAs methods and report the results on NYUv2 and Pascal in \cref{tab:nyuv2sota} and \cref{tab:pascalsota}, respectively.

On NYUv2, integrating our method with state-of-the-art MTL methods consistently improves their MTL performance by 
average over 1.7. Notably, our approach leads to comprehensive improvements on RADIO~\cite{ranzinger2024radio} and DINOv3~\cite{simeoni2025dinov3} across all tasks, and surpasses SAK~\cite{lu2024swiss} in three out of four tasks, with comparable segmentation result. 
Moreover, our CvM demonstrates clear benefits in geometry-intensive tasks such as depth estimation. 
Across three MTL methods, our method improves depth by 4.29\% on average (\eg, from 0.4113 (Dinov3) to 0.3909 (Ours)) and boosts boundary detection F-score by 1.2 on average (\eg,  Ours vs Dinov3: 81.58 vs 80.10). 
Our method with Dinov3 achieves a new SotA across all tasks on NYUv2.

On PASCAL‑Context, our method also brings consistent improvements over SotA MTL approaches. By integrating CvM into RADIO~\cite{ranzinger2024radio}, SAK \cite{lu2024swiss}, and DINOv3~\cite{simeoni2025dinov3}, our method yields gains across all tasks and boosts MTL performance. The results show that this hybrid design is beneficial: despite the absence of multi‑view inputs, our CvM still helps to prevent MTL encoder from capturing noisy and spurious view correlations between identical views and further improves the MTL performance. 

{
\setlength{\tabcolsep}{3.2pt} 
\begin{table}[t]
	\centering
	
    \resizebox{0.47\textwidth}{!}
    {
		\begin{tabular}{lcccccccccc}

		    \toprule
		     \multirow{2}{*}{Method}  & Seg.  & PartSeg & Sal & Normal & Boundary & \multirow{2}{*}{$\Delta$MTL $\uparrow$} \\
              &  (mIoU) $\uparrow$ & (mIoU) $\uparrow$ & (maxF) $\uparrow$ & (mErr) $\downarrow$ & (odsF) $\uparrow$ & \\
            \midrule
            STL &  81.61 & 72.77 & 83.80 & 13.87 & 75.24 & 0.00 \\
            MTL & 79.26 & 68.28 & 84.16 & 14.06 & 71.59 & -2.97 \\
            \midrule
            TaskExpert~\cite{ye2023taskexpert} & 80.64 & 69.42 & 84.87 & 13.56 & 73.30 & -0.97\\
            BFCI~\cite{zhang2023rethinking} & 80.64 & 70.06 & 84.64 & 13.82 & 72.96 & -1.32\\
            TSP~\cite{wang2024tsp} & 81.48 & 70.64 & 84.86 & 13.69 & 74.80 & -0.22\\
            MLoRE~\cite{yang2024multi} & 81.41 & 70.52 & 84.90 & 13.51 & 75.42 & 0.16\\
            InvPT~\citep{ye2022inverted}        & 79.03 & 67.61 & 84.81 & 14.15 & 73.00 & -2.81 \\
            3DMTL~\citep{li2023multi} &  79.53 & 69.12 & 84.94 & 13.53 & 74.00 & -1.08\\
            \midrule
            RADIO~\citep{ranzinger2024radio} & 81.11 & 71.50 & 85.17 & 13.49 & 74.80 & 0.29 \\
            \rowcolor{LightRed} {\bf Ours} & {\bf 81.18} & {\bf 71.75} & {\bf 85.26} & {\bf 13.42} & {\bf 76.95} & {\bf 1.07} \\
            \midrule
            SAK~\citep{lu2024swiss} &  84.01 & 76.99 & 84.65 & 13.82 & 76.27 & 2.30 \\
            \rowcolor{LightRed} {\bf Ours} & {\bf 84.41} & {\bf 77.68} & {\bf 84.83} & {\bf 13.61} & {\bf 76.79} & {\bf 3.07} \\
            \midrule
            DINOv3~\citep{simeoni2025dinov3} & 84.07 & 77.29 & 84.40 & 13.70 & 76.30 & 2.52 \\
            \rowcolor{LightRed} {\bf Ours} & {\bf 84.56} & {\bf 77.97} & {\bf 84.56} & {\bf 13.66} & {\bf 79.29} & {\bf 3.71} \\
			\bottomrule
		\end{tabular}%
			}
		\vspace{-0.25cm}
		\caption{Quantitative comparison of our method to the SotA methods on PASCAL-Context dataset. $\Delta$MTL is computed using single-task learning ``STL'' as baseline.}
        \vspace{-0.3cm}
		\label{tab:pascalsota}
\end{table}%
}

\paragraph{Training Cost.}
Our CvM introduces approximately 5M additional parameters, making it a lightweight component relative to the overall size of typical MTL encoders, \eg, MTL models like RADIO~\cite{ranzinger2024radio}, SAK~\cite{lu2024swiss}, and DINOv3~\cite{simeoni2025dinov3} typically contain 300–350M parameters. Our proposed CvM accounts for only 1.5\% of the total parameter count of MTL encoder. 

\subsection{Ablation study}\label{sec:exp:abla}
Here we conduct an in-depth analysis of our method to validate the effectiveness of our CvM. All experimental analyses are performed on NYUv2 dataset. Video frames in NYUv2 are used during training. Please refer to supplementary for more detailed results of \cref{tab:nyuv23Ddesign,tab:nyuv2depthcandidates,tab:nyuv2differentviews}

\paragraph{Cost Volume \& Cross-view Features.}
We aim to examine the contributions of the cost volume $\bC_i$ and the cross‑view enhanced features $\bF_i$ in our method. 
We start with our method without both $\bC_i$ and $\bF_i$, resulting in a standard MTL baseline ``Ours \emph{w/o CV \& CF}''. We then add only the cost volume, \ie, ``Ours \emph{w/o CF}'' to verify the effectiveness of cost volume. Based on ``Ours w/o CF'', we further add the cross‑view features, which is our full model ``Ours''. 

{
\setlength{\tabcolsep}{3.2pt} 
\begin{table}[t]
	\centering
	
    \resizebox{0.47\textwidth}{!}
    {
		\begin{tabular}{lccccccccc}

		    \toprule
		     Method 
             & \makecell{Seg.\\(mIoU) $\uparrow$} 
             & \makecell{Depth\\(RMSE) $\downarrow$} 
             & \makecell{Normal\\(mErr) $\downarrow$} 
             & \makecell{Boundary\\(odsF) $\uparrow$}  
             & \makecell{$\Delta$MTL $\uparrow$} \\
		    \midrule
            Ours \emph{w/o CV \& CF} & 64.03 & 0.3954 & 15.35 & 80.52 & 17.57 \\ 
            \rowcolor{LightRed} Ours \emph{w/o CF} & 64.86 & 0.3853 & {\bf 15.33} & 81.18 & 18.65 \\ 
           \rowcolor{LightRed} Ours & {\bf 65.27} & {\bf 0.3836} & 15.35 & {\bf 81.69} & {\bf 19.05}\\ 
			\bottomrule
		\end{tabular}%
			}
		\vspace{-0.25cm}
		\caption{Ablation study on cross-view module. ``Ours \emph{w/o CV \& CF}'' is the baseline without our cross-view module. ``Ours \emph{w/o CF}'' indicate our method that only uses cost volume. ``Ours'' is our full model that uses cost volume and cross-view feature. $\Delta$MTL is computed using ``STL'' in \cref{tab:nyuv2sota} as baseline.}
        \vspace{-0.3cm}
		\label{tab:nyuv2component}
\end{table}%
}

\begin{table*}[t]
\centering
\setlength{\tabcolsep}{4pt}

\begin{minipage}{0.41\textwidth}
\centering
\resizebox{\textwidth}{!}
    {
		\begin{tabular}{cccc>{\columncolor{LightRed}}cccccc}

		    \toprule
		    \begin{tabular}[c]{@{}c@{}} Spatial-aware \\ Encoder\end{tabular} & \begin{tabular}[c]{@{}c@{}} MTL \\ Encoder\end{tabular} & \begin{tabular}[c]{@{}c@{}} MTL Encoder \\ + LoRA~\cite{hu2022lora} \end{tabular} & \begin{tabular}[c]{@{}c@{}} MTL Encoder \\ + Adapter~\cite{lu2024swiss} \end{tabular} & Ours \\             
		    \midrule
             $\Delta$MTL $\uparrow$ & 18.84 & 18.87 & 18.98 & {\bf 19.05} \\
			\bottomrule
		\end{tabular}%
			}
		\vspace{-0.3cm}
		\caption{Comparisons of various spatial-aware features methods in cross-view module on NYUv2. $\Delta$MTL is computed using ``STL'' in \cref{tab:nyuv2sota} as baseline.}
		\label{tab:nyuv23Ddesign}
\end{minipage}
\hfill
\begin{minipage}{0.32\textwidth}
\centering
\resizebox{\textwidth}{!}
    {
		\begin{tabular}{lc>{\columncolor{LightRed}}cccccccc}

		    \toprule
		    \begin{tabular}[c]{@{}c@{}} Number of \\ Depth \end{tabular} & 96 & 128 & 256 & 384 & 512 \\
		    \midrule
             $\Delta$MTL $\uparrow$ &  18.88 & 19.05 & 18.62 & 18.46 & {\bf 19.25} \\

			\bottomrule
		\end{tabular}%
			}
		\vspace{-0.3cm}
		\caption{Comparisons of various number of depth candidates on NYUv2. $\Delta$MTL is computed using ``STL'' in \cref{tab:nyuv2sota} as baseline.}
		\label{tab:nyuv2depthcandidates}
\end{minipage}
\hfill
\begin{minipage}{0.235\textwidth}
\centering
\resizebox{\textwidth}{!}
    {
		\begin{tabular}{ccc>{\columncolor{LightRed}}ccccccc}

		    \toprule
		     \begin{tabular}[c]{@{}c@{}} Number of \\ Views \end{tabular} & 4 & 3 & 2 \\
             \midrule
             $\Delta$MTL $\uparrow$ & 18.63 & {\bf 19.20} & 19.05\\
			\bottomrule
		\end{tabular}%
			}
		\vspace{-0.3cm}
		\caption{Results of various number of views. $\Delta$MTL is computed w.r.t. ``STL'' in \cref{tab:nyuv2sota}.}
		\label{tab:nyuv2differentviews}
\end{minipage}
\vspace{-0.3cm}
\end{table*}

\begin{figure*}[t]
    \centering
    \includegraphics[width=0.95\textwidth]{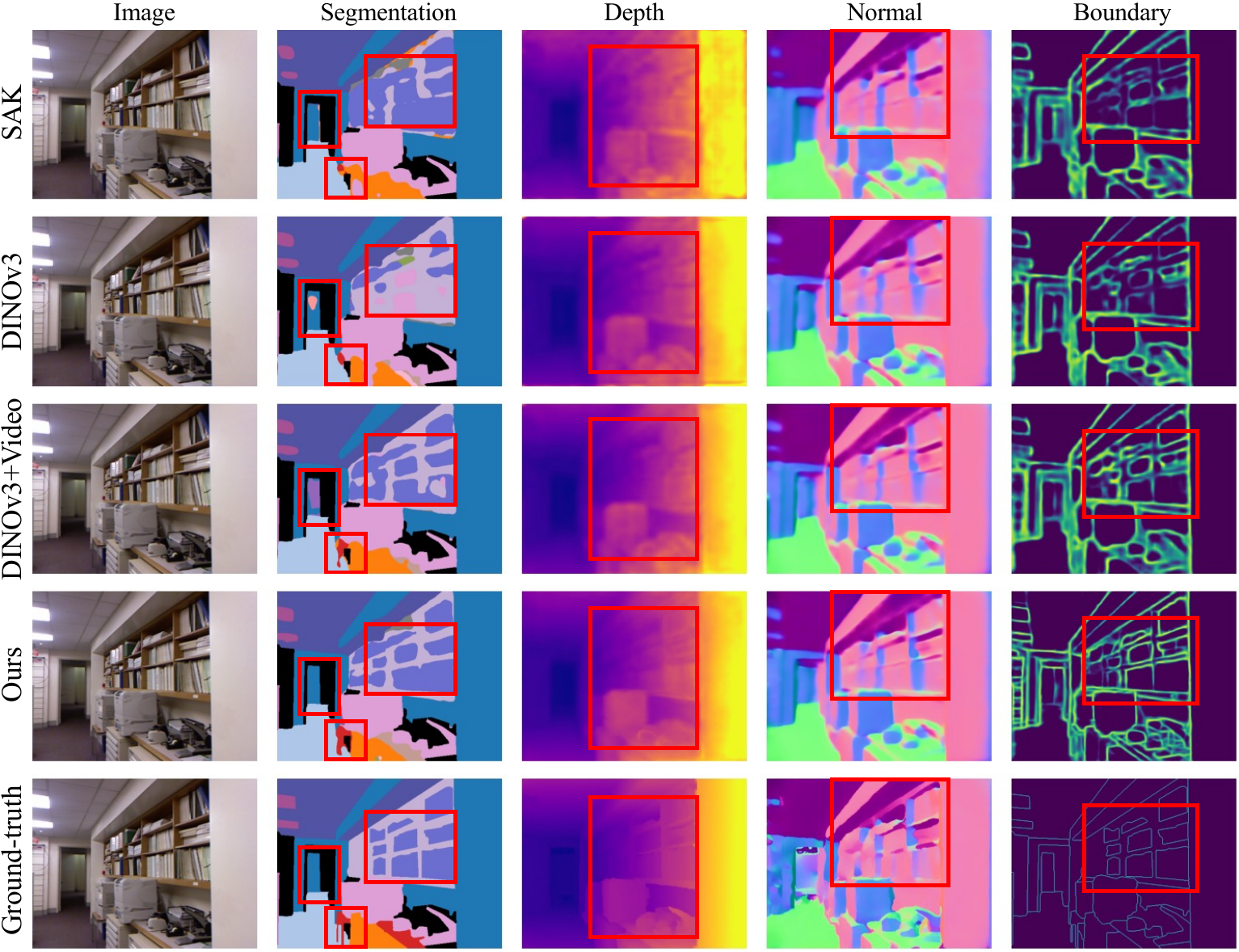}
    \vspace{-0.3cm}
    \caption{\textbf{Qualitative Comparisons on NYUv2.} The first column shows the RGB image, and the remaining columns display either the ground-truth or model predictions. The last row shows the ground-truth of four tasks. The first to the fourth row shows the predictions of SAK, Dinov3, Dinov3 trained with videos as multi-view data, and our method, respectively.}

    \label{fig:visual}
\end{figure*}

The results are shown in \cref{tab:nyuv2component}, where we can see that the cost volume alone effectively supplements the MTL encoder with cross‑view geometric cues and improves the MTL performance ($\Delta$MTL) by an average of over 1\%. When combined with the cross‑view enhanced features, our model achieves a further boost in performance, indicating that the two components are complementary. These findings validate the effectiveness of our design and demonstrate that injecting cross‑view information into a standard MTL encoder is beneficial for learning geometry‑aware representations that enhance performance across multiple tasks.

\paragraph{Extracting spatial-aware features} plays a crucial role for subsequent cost volume construction and 3D-aware MTL. Apart from our design, we also consider three other methods and report results in \cref{tab:nyuv23Ddesign}:
(1) ``MTL encoder'' uses the feature from the MTL encoder as spatial-aware feature.
(2) ``MTL encoder + LoRA~\cite{hu2022lora}'' attaches low-rank adapters (LoRA)~\cite{hu2022lora} (rank and $\alpha$ is set to 16, scaling factor $s$=1) into the MTL encoder to adapt the features from MTL encoder as spatial-aware features.
(3) ``MTL encoder + Adapter~\cite{lu2024swiss}'' appends adapters from SAK~\cite{lu2024swiss} to the MTL encoder for encoding spatial-aware features. 
As shown in \cref{tab:nyuv23Ddesign}, our design achieves the highest MTL performance. We attribute this to the strong inductive bias of CNNs in modeling spatial structures, resulting in higher-quality spatial-aware features. 
Beyond performance advantages, using a lightweight independent spatial-aware encoder instead of modifying the MTL encoder offers a non-intrusive mechanism to supply 3D-aware features, making our method architecture-agnostic and easily integrable with various MTL backbones.

\paragraph{Number of Depth Candidates $L$} can affect the reconstruction of cost volume,
and we experiment with 128, 256, 384, and 512 candidates while keeping the depth range fixed to investigate the effect of $L$. As shown in \cref{tab:nyuv2depthcandidates}, increasing the number of candidates to 512 improves performance.
However, using 512 depth candidates inevitably increases the computational cost significantly. For better trade-off between efficiency and performance, we use 128 depth candidates. This also aligns with the choices in 3D reconstruction pipelines such as MVSplat \cite{chen2024mvsplat} and DepthSplat \cite{xu2025depthsplat}.

\paragraph{Number of Views.} We use one neighbour view (V=2) while our method supports more views. We performed experiments with $V=2,3,4$ and results are reported in \cref{tab:nyuv2differentviews}. We can see that increasing the number of views can help learning better geometric shapes and improve performance, while using $V=2$ is sufficient and efficient. 

\subsection{Qualitative Results}
We visualize task predictions for four methods: SAK \cite{lu2024swiss}, DINOv3~\cite{simeoni2025dinov3}, DINOv3 trained with multi-view video data, and Ours. 
As shown in \cref{fig:visual}, SAK and single-view DINOv3 mis-segment the bookshelf and table-leg areas and produce blurred depth and noisy normals. DINOv3 trained with multi-view data improves the prediction but it still fails to recover fine geometry and boundaries. Our method can be observed to estimate more accurate predictions, yielding accurate segmentation of thin structures, sharper depth discontinuities, more stable normals, and clearer boundaries. These results strongly verify that geometric information is crucial for comprehensive scene understanding, and our method is capable of injecting geometric consistency into MTL methods.

\section{Conclusion and Future Work}\label{sec:con}
In this paper, we demonstrate that cross-view consistency provides crucial geometric cues for multi-task dense prediction in scene understanding across several benchmarks. We introduce a Cross-view Module (CvM) for MTL that estimates cross-view correlations through a spatial-aware encoder with a multi-view transformer and cost-volume construction. 
Through extensive experiments, we have demonstrated that CvM can seamlessly integrate into existing multi-task learning architectures and supports both single- and multi-view input. 
Despite its effectiveness, our method has limitations: it is designed for static scenes, whereas dynamic environments with moving objects or camera motion introduce additional challenges. 
Future work will explore more efficient multi-view modeling and motion-aware extensions to better handle dynamic scenes.
\section*{Acknowledgement}
We thank Dima Damen, Xiang Li, Yuedong Chen and Haofei Xu for their valuable feedback and comments. This work was supported in part by access to the Isambard-AI system through the UKRI Gateway Scheme.

{
    \small
    \bibliographystyle{ieeenat_fullname}
    \bibliography{main}
}
\clearpage
\appendix

\renewcommand{\thesection}{A\arabic{section}}
\renewcommand{\thesubsection}{A\arabic{section}.\arabic{subsection}}
\renewcommand{\thefigure}{A\arabic{figure}}
\renewcommand{\thetable}{A\arabic{table}}


\section{More Details}
\subsection{Implementation Details}
We apply our method to different state-of-the-art MTL methods, including SAK~\cite{lu2024swiss}, Radio~\cite{ranzinger2024radio} from Lu \etal~\cite{lu2024swiss} and DINOv3~\cite{simeoni2025dinov3}, by attaching the cross-view module (CvM) to their encoder. 
For all experiments, we follow identical training and evaluation protocal of prior work~\cite{lu2024swiss}.
We implement our model in PyTorch \cite{paszke2019pytorch} and use AdamW \cite{loshchilov2017decoupled} as optimizer with a learning rate of $2 \times 10^{-5}$ and a weight decay rate of $1 \times 10^{-6}$. Polynomial learning rate scheduler is used to dynamically adjust the learning rate. We use a batch size of 4 and train each model for 40000 steps.
The image size is $448 \times 576$ for NYUv2 and $512 \times 512$ for PASCAL-Context. We use the same loss functions and loss weights as in Lu \etal~\cite{lu2024swiss,ye2022inverted,vandenhende2021multi}: cross-entroy loss for segmentation, human parsing, saliency and boundary detection, L1 loss for depth and normal estimation. 
Across all experiments, we use ViT-L as backbone for MTL encoder, and utilize multi-scale features for vision transformer, i.e., intermediate features from layer 5, 12, 18, 24. More details about the design of CvM is presented in \cref{suppsec:cvm_design}.

\subsection{Details for CvM}\label{suppsec:cvm_design}
In our CvM, we implement a ResNet-style \cite{he2016deep} convolutional network as the spatial-aware encoder to extract geometry-sensitive features from multi-view RGB inputs. The encoder consists of three residual blocks, progressively reducing spatial resolution while increasing channel dimensions, ultimately producing a 128-dimensional feature map at $1/8$ the input resolution. The spatial features are then fed into the multi-view transformer module.

The multi-view transformer comprises six layers of self-attention and cross-view attention, each employing the Swin Transformer’s \cite{liu2021swin} window-based attention mechanism to preserve local context while enabling efficient cross-view communication. To better align the cross-view enhanced features with the MTL feature space, we remove the output normalization of the final cross-attention layer and instead append a lightweight SwiGLU-based \cite{shazeer2020glu} adapter module, which consists of a gated MLP layer. 
The resulting cross-view enhanced features are subsequently aligned using the camera intrinsics and relative poses between views to construct a cost volume. 
Specifically, given a set of depth candidates ${d_1, \dots, d_L}$ sampled in inverse-depth space over a range of $0.0001$ to $10$ meters, we follow a differentiable feature warping strategy to reproject the features from neighboring views onto the coordinate frame of the reference view. Concretely, for each pixel location in the neighboring view, we back-project it into a 3D point at each hypothesized depth using the its camera intrinsics. These 3D points are then transformed into the coordinate system of the reference view using the relative camera pose. The transformed 3D points are subsequently reprojected into the reference image plane using its intrinsics, yielding a dense sampling grid across depth planes. The resulting warped features are used to construct a volumetric cost volume, which encodes the view-wise matching information across different depth planes and serves as a strong 3D-aware cue for subsequent MTL prediction.

Then, both the cross-view enhanced features and cost volume are then upsampled by a learned 4$\times$ upsampling module. These upsampled features are concatenated with the multi-scale features from the MTL encoder and fused within the task-specific decoder heads. Finally, a linear layer takes the fused 3D-aware multi-task feature as input and regresses the MTL predictions for each task.

\section{Additional Results}
\subsection{Detailed Results for Ablation Study}
The detailed task-specific results for our ablation study on \textit{extracting spatial-aware features} (Tab. 5 in the main paper), \textit{number of Depth Candidates $L$} (Tab. 6 in the main paper) and \textit{number of views} (Tab. 7 in the main paper) are presented in the following \cref{supptab:nyuv23Ddesign_ab}, \cref{supptab:nyuv2depthcandidates_ab} and \cref{supptab:nyuv2differentviews_ab}, respectively.

{
\setlength{\tabcolsep}{3.2pt} 
\begin{table}[htbp]
	\centering
	
    \resizebox{0.47\textwidth}{!}
    {
		\begin{tabular}{lccccccccc}

		    \toprule
             \makecell[l]{Spatial-aware \\ Encoder} 
             & \makecell{Seg.\\(mIoU) $\uparrow$} 
             & \makecell{Depth\\(RMSE) $\downarrow$} 
             & \makecell{Normal\\(mErr) $\downarrow$} 
             & \makecell{Boundary\\(odsF) $\uparrow$}  
             & \makecell{$\Delta$MTL $\uparrow$} \\
		    \midrule
            MTL Encoder & {\bf 65.34} & 0.3847 & 15.52 & 80.54 & 18.84 \\
            \makecell[l]{MTL Encoder \\+ LoRA~\cite{hu2022lora}} & 64.83 & {\bf 0.3760} & 15.40 & 80.89 & 18.87\\
            \makecell[l]{MTL Encoder \\ + Adapter~\cite{lu2024swiss}} & {\bf 65.34} & 0.3814 & {\bf 15.30} & 80.85 & 18.98\\
            \rowcolor{LightRed} Ours & 65.27 & 0.3836 & 15.35 & {\bf 81.69} & {\bf 19.05} \\
			\bottomrule
		\end{tabular}%
			}
		\vspace{-0.25cm}
		\caption{Detailed MTL performance for comparisons of various spatial-aware features methods in cross-view module on NYUv2. $\Delta$MTL is computed using single-task learning ``STL'' in Tab. 2 in the main paper as baseline.}
		\label{supptab:nyuv23Ddesign_ab}
\end{table}%
}

\paragraph{Detailed Results for Ablation on Extracting Spatial-aware Features.}
As shown in \cref{supptab:nyuv23Ddesign_ab}, although different designs for extracting spatial-aware features exhibit varying performance across tasks, our CNN-based design achieves the highest overall MTL performance. This architecture-agnostic and non-intrusive design also makes CvM readily applicable to a wider range of MTL encoders, highlighting its practical flexibility and generalizability.

{
\setlength{\tabcolsep}{3.2pt} 
\begin{table}[htbp]
	\centering
    \resizebox{0.47\textwidth}{!}
    {
		\begin{tabular}{lccccccccc}

		    \toprule
             \makecell[l]{\#Depth \\Candidates} 
             & \makecell{Seg.\\(mIoU) $\uparrow$} 
             & \makecell{Depth\\(RMSE) $\downarrow$} 
             & \makecell{Normal\\(mErr) $\downarrow$} 
             & \makecell{Boundary\\(odsF) $\uparrow$}  
             & \makecell{$\Delta$MTL $\uparrow$} \\
		    \midrule
            96 & 64.84 & 0.3827 & 15.32 & 81.54 & 18.88\\

            \rowcolor{LightRed} 128 & 65.27 & 0.3836 & 15.35 & {\bf 81.69} & 19.05\\

            256 & 64.21 & 0.3832 & {\bf 15.29} & 81.58 & 18.62  \\

            384 & 64.08 & 0.3835 & 15.34 & 81.51 & 18.46 \\

            \rowcolor{LightRed}  512 & {\bf 65.28} & {\bf 0.3778} & 15.37 & 81.57 & {\bf 19.25} \\

			\bottomrule
		\end{tabular}%
			}
		\vspace{-0.25cm}
		\caption{Detailed MTL performance for comparisons of various number of depth candidates on NYUv2. $\Delta$MTL is computed using single-task learning ``STL'' in Tab. 2 in the main paper as baseline.}
        \vspace{-0.25cm}
		\label{supptab:nyuv2depthcandidates_ab}
\end{table}%
}

\paragraph{Detailed Results for Ablation on Number of Depth Candidates.}
In \cref{supptab:nyuv2depthcandidates_ab}, increasing the number of depth candidates to 512 significantly improves depth estimation performance, reducing the RMSE from $0.3836$ (with 128 candidates) to $0.3778$, while maintaining comparable performance on the other tasks. However, such a fine-grained discretization of the depth range introduces substantial computational overhead due to the increased cost of feature warping during cost volume construction. Considering the trade-off between performance and efficiency, setting $L=128$ offers an optimal balance, which aligns with commonly adopted settings in recent 3D reconstruction pipelines such as MVSplat \cite{chen2024mvsplat} and DepthSplat \cite{xu2025depthsplat}.

\paragraph{Detailed Results for Ablation on Number of Views.}
In \cref{supptab:nyuv2differentviews_ab}, we observe that increasing the number of input views from 2 to 3 leads to improved overall MTL performance. However, further increasing the number of views to 4 results in a slight performance drop. We attribute this to the increased complexity of learning cross-view correlations from multiple views, which, when combined with the inherent challenge of balancing multiple tasks in MTL, may hinder effective optimization. Moreover, as NYUv2 video sequences lack ground-truth camera poses, we rely on poses estimated via COLMAP \cite{schoenberger2016sfm}, which may introduce noise. Using more views could amplify such pose estimation errors, negatively impacting the quality of learned geometric features. Future work may explore pose-free alternatives or incorporate view-relative pose prediction to enable more robust multi-view training.

{
\setlength{\tabcolsep}{3.2pt} 
\vspace{-0.15cm}
\begin{table}[htbp]
	\centering
    \resizebox{0.47\textwidth}{!}
    {
		\begin{tabular}{lccccccccc}

		    \toprule
              \makecell[l]{Input \\Views}
             & \makecell{Seg.\\(mIoU) $\uparrow$} 
             & \makecell{Depth\\(RMSE) $\downarrow$} 
             & \makecell{Normal\\(mErr) $\downarrow$} 
             & \makecell{Boundary\\(odsF) $\uparrow$}  
             & \makecell{$\Delta$MTL $\uparrow$} \\
		    \midrule
            \rowcolor{LightRed} {2} & {\bf 65.27} & { 0.3836} & 15.35 & {81.69} & { 19.05} \\ 
            {\bf 3} & 65.17 & {\bf 0.3827} & {\bf 15.24} & {\bf 81.72} & {\bf 19.20} \\
            4 & 64.97 & 0.3913 & 15.28 & 81.61 & 18.63 \\

			\bottomrule
		\end{tabular}%
			}
		\vspace{-0.25cm}
		\caption{Detailed MTL performance for results of various number of views. $\Delta$MTL is computed using single-task learning ``STL'' in Tab. 2 in the main paper as baseline.}
        \vspace{-0.45cm}
		\label{supptab:nyuv2differentviews_ab}
\end{table}%
}

\subsection{Depth Range for Pascal Dataset}
Unlike the NYUv2 \cite{silberman2012indoor} dataset, which provides depth labels and has a well-defined depth range of 0–10 meters, the PASCAL-Context dataset does not include ground-truth depth. As a result, the appropriate depth range for constructing the cost volume remains uncertain. Given that PASCAL-Context contains a diverse set of indoor and outdoor scenes, we explore different candidate depth ranges while keeping the number of candidates fixed at $L=128$. Specifically, we evaluate three configurations: $\left(0.0001, 10.0\right)$ (following the NYUv2 setting), $\left(0.0001, 50.0\right)$, and $\left(0.0001, 100.0\right)$. The results are reported in \cref{supptab:pascaldepthrange}. Among all settings, reusing the NYUv2 configuration $\left(0.0001, 10.0\right)$ yields slightly better MTL performance. However, the differences across the three depth range configurations are marginal, indicating that our method is not particularly sensitive to the choice of depth range on the PASCAL-Context dataset. Therefore, we adopt the NYUv2 configuration $\left(0.0001, 10.0\right)$ for consistency with the experiments conducted on NYUv2.
\begin{table}[htbp]
	\centering
    \resizebox{0.47\textwidth}{!}
    {
		\begin{tabular}{lccccccccc}

		    \toprule
		     \multirow{2}{*}{Depth Range} &  Seg.  & PartSeg & Sal & Normal & Boundary & \multirow{2}{*}{$\Delta$MTL $\uparrow$} \\
              &  (mIoU) $\uparrow$ & (mIoU) $\uparrow$ & (maxF) $\uparrow$ & (mErr) $\downarrow$ & (odsF) $\uparrow$ &\\
		    \midrule

            DINOv3 baseline & 84.07 & 77.29 & 84.40 & 13.70 & 76.30 & 2.52 \\

            \rowcolor{LightRed} \textbf{$\left(0.0001, 10.0\right)$ (Ours)} &  84.56 & {\bf 77.97} & {\bf 84.56} & {\bf 13.66} & 79.29 & 3.71 \\

            $\left(0.0001, 50.0\right)$ & {\bf 84.58} & 77.89 & 84.52 & 13.67 & 79.28 & 3.67 \\
            
            $\left(0.0001, 100.0\right)$ & 84.50 & 77.86 & 84.50 & {\bf 13.66} & {\bf 79.32} & 3.66 \\

			\bottomrule
		\end{tabular}%
			}
		\vspace{-0.25cm}
		\caption{Ablation on depth range for constructing depth candidates on PASCAL-Context. $\Delta$MTL is computed using single-task learning ``STL'' in Tab. 3 in the main paper as baseline.}
		\label{supptab:pascaldepthrange}
\end{table}%

\subsection{Results with ViT-B Backbones}

In this section, we provide results with ViT-B backbones to evaluate the generality of our method across different backbone capacities and to complement the main results reported with ViT-L. 

\paragraph{MTL with Multiple Views.}
We first conduct multi-view MTL experiments following the setup of Tab. 1 in the main paper. Results are presented in \cref{supptab:nyuv2mv}. After reducing the number of parameters in the MTL encoder, it becomes increasingly difficult for the model to directly learn effective cross-view correlations and 3D awareness by simply introducing multi-view data during training. Although this strategy still improves overall MTL performance, it proves inefficient for injecting 3D priors and shows limited benefits across individual tasks.

In contrast, our CvM makes more effective use of the multi-view data, enabling efficient injection of 3D awareness into the MTL model. Specifically, our CvM leads to an MTL performance gain of +2.31 and +3.19 for SAK and DINOv3, respectively, compared to their baselines trained without video data. Notably, when applied to DINOv3, our method not only surpasses its multi-view trained counterpart but also outperforms 3DMTL \cite{li2023multi} across all tasks. These results further confirm that CvM learns cross-view correlations more effectively and consistently enhances the performance of MTL models by introducing 3D geometric awareness.

{
\setlength{\tabcolsep}{3.2pt} 
\begin{table}[htbp]
	\centering
	
    \resizebox{0.47\textwidth}{!}
    {
		\begin{tabular}{lccccccccc}

		    \toprule
		     Method 
             & \makecell{Seg.\\(mIoU) $\uparrow$} 
             & \makecell{Depth\\(RMSE) $\downarrow$} 
             & \makecell{Normal\\(mErr) $\downarrow$} 
             & \makecell{Boundary\\(odsF) $\uparrow$}  
             & \makecell{$\Delta$MTL $\uparrow$} \\
            \midrule
            SAK~\citep{lu2024swiss} \emph{w/o video} & {\bf 59.93} & 0.4942 & 17.60 & 78.60 & 0.00 \\
            SAK~\citep{lu2024swiss} & 59.41 & 0.4718 & 17.65 & 78.38 & 0.78 \\ 

            \rowcolor{LightRed} {\bf Ours} & {58.97} & {\bf 0.4534} & {\bf 17.40} & {\bf 79.74} & {\bf 2.31}\\ 
            \midrule
            
            DINOv3~\citep{simeoni2025dinov3}~ \emph{w/o video} & 59.73 & 0.4650 & 16.80 & 78.53 & 0.00 \\
            DINOv3~\citep{simeoni2025dinov3} & 59.72 & 0.4450 & 16.90 & 78.40 & 0.88\\ 
            3DMTL$^{*}$ & 59.65 & 0.4403 & 16.72 & 78.72 & 1.47 \\
            \rowcolor{LightRed} {\bf Ours} & {\bf 60.74} & {\bf 0.4263} & {\bf 16.66} & {\bf 80.03} & {\bf 3.19}\\ 
			\bottomrule
		\end{tabular}%
			}
		\vspace{-0.25cm}
		\caption{Quantitative comparison of our method with ViT-B backbone on NYUv2 dataset + extra video frames with multiple views. $^{*}$: We reproduce 3DMTL~\cite{li2023multi} with DINOv3 backbone. $\Delta$MTL is computed using ``SAK~\cite{lu2024swiss} \emph{w/o video}'' and ``DINOv3~\cite{simeoni2025dinov3} \emph{w/o video}'' as baseline, respectively.}
		\label{supptab:nyuv2mv}
\end{table}%
}

\paragraph{Comparison with SotAs.}
When integrating our CvM into state-of-the-art MTL frameworks with ViT-B backbones, following the same setting of Tab. 2 and Tab. 3 in the main paper, we observe a similar trend of performance improvement as with ViT-L. On the NYUv2 dataset, our method consistently enhances the performance of both RADIO~\cite{ranzinger2024radio} and DINOv3~\cite{simeoni2025dinov3} across all tasks. It also surpasses SAK~\cite{lu2024swiss} on three out of four tasks, while achieving comparable segmentation performance. On the PASCAL-Context dataset, our CvM again delivers comprehensive gains for all three MTL encoders, demonstrating a similar trend to the ViT-L backbone results. Detailed comparisons are provided in \cref{supptab:nyuv2sota} and \cref{supptab:pascalsota}.

{
\setlength{\tabcolsep}{3.2pt} 
\begin{table}[t]
    \centering
	
    \resizebox{0.47\textwidth}{!}
    {
		\begin{tabular}{lcccccccccc}

		    \toprule
		     \multirow{2}{*}{Method} &  Seg. & Depth & Normal & Boundary & \multirow{2}{*}{$\Delta$MTL $\uparrow$}  \\
             & (mIoU) $\uparrow$ & (RMSE) $\downarrow$ & (mErr) $\downarrow$ & (odsF) $\uparrow$ & \\
		    \midrule
            STL  & 51.15 & 0.5792 & 19.77 & 77.35 & 0.00 \\
            MTL &  49.27 & 0.5823 & 19.92 & 75.88 & -1.72 \\
            \midrule
            BFCI~\cite{zhang2023rethinking} &  51.14 & 0.5186 & 18.92 & 77.98 & 3.89 \\
            TSP~\cite{wang2024tsp}  & 51.22 & 0.5301 & 18.78 & 76.90 & 3.26 \\
            InvPT~\citep{ye2022inverted}  & 50.30 & 0.5367 & 19.00 & 77.60 & 2.47 \\
            \midrule RADIO~\citep{ranzinger2024radio}  & 55.03 & 0.5186 & 18.49 & 77.97 & 6.33 \\
            \rowcolor{LightRed}{\bf Ours}  & {\bf 55.96} & {\bf 0.4970} & {\bf 18.36} & {\bf 79.35} & {\bf 8.32} \\
            \midrule
            SAK~\citep{lu2024swiss}  & {\bf 59.93} & 0.4942 & 17.60 & 78.60 & 11.11 \\
            \rowcolor{LightRed} {\bf Ours} & 59.60 & {\bf 0.4535} & {\bf 17.34} & {\bf 79.95} & {\bf 13.47} \\
            \midrule
            DINOv3~\citep{simeoni2025dinov3} & 59.73 & 0.4650 & 16.80 & 78.53 & 13.26 \\
            \rowcolor{LightRed} {\bf Ours} & {\bf 60.61} & {\bf 0.4376} & {\bf 16.63} & {\bf 80.38} & {\bf 15.69} \\ 
			\bottomrule
		\end{tabular}%
			}
		\vspace{-0.25cm}
		\caption{Quantitative comparison of our method with ViT-B backbone to the SotA methods on NYUv2 dataset. $\Delta$MTL is computed using single-task learning ``STL'' as baseline. }
		\label{supptab:nyuv2sota}
\end{table}%
}

{
\setlength{\tabcolsep}{3.2pt} 
\begin{table}[t]
	\centering
	
    \resizebox{0.47\textwidth}{!}
    {
		\begin{tabular}{lcccccccccc}

		    \toprule
		     \multirow{2}{*}{Method}  & Seg.  & PartSeg & Sal & Normal & Boundary & \multirow{2}{*}{$\Delta$MTL $\uparrow$} \\
              &  (mIoU) $\uparrow$ & (mIoU) $\uparrow$ & (maxF) $\uparrow$ & (mErr) $\downarrow$ & (odsF) $\uparrow$ & \\
            \midrule
            STL &  80.25 & 70.54 & 84.54 & 13.57 & 74.22 & 0.00 \\
            MTL & 76.76 & 65.26 & 84.39 & 13.98 & 70.37 & -4.04 \\
            \midrule
            TaskExpert~\cite{ye2023taskexpert} & 78.45 & 67.38 & 84.96 & 13.55 & 72.30 & -1.73\\
            BFCI~\cite{zhang2023rethinking} & 77.98 & 68.19 & 85.06 & 13.48 & 72.98 & -1.31\\
            MLoRE~\cite{yang2024multi} & 79.26 & 67.82 & 85.31 & 13.65 & 74.69 & -0.83\\
            InvPT~\citep{ye2022inverted}  & 77.33 & 66.62 & 85.14 & 13.78 & 73.20 & -2.28 \\
            \midrule
            RADIO~\citep{ranzinger2024radio} & 78.06 & 68.13 & 85.18 & 13.59 & 72.64 & -1.53 \\
            \rowcolor{LightRed} {\bf Ours} & {\bf 78.21} & {\bf 69.20} & {\bf 85.20} & {\bf 13.50} & {\bf 75.82} & {\bf -0.20} \\
            \midrule
            SAK~\citep{lu2024swiss} &  81.88 & 74.30 & 84.79 & 14.02 & 74.09 & 0.83 \\
            \rowcolor{LightRed} {\bf Ours} & {\bf 81.94} & {\bf 75.22} & {\bf 84.90} & {\bf 13.72} & {\bf 77.76} & {\bf 2.57} \\
            \midrule
            DINOv3~\citep{simeoni2025dinov3} & 81.46 & 74.11 & 84.71 & 13.81 & 73.94 & 2.52 \\
            \rowcolor{LightRed} {\bf Ours} & {\bf 82.10} & {\bf 75.06} & {\bf 85.18} & {\bf 13.67} & {\bf 77.64} & {\bf 2.69} \\
			\bottomrule
		\end{tabular}%
			}
		\vspace{-0.25cm}
		\caption{Quantitative comparison of our method with ViT-B backbone to the SotA methods on PASCAL-Context dataset. $\Delta$MTL is computed using single-task learning ``STL'' as baseline. }
        \vspace{-0.3cm}
		\label{supptab:pascalsota}
\end{table}%
}

\section{Computational Cost Analysis}
We analyze the computational overhead introduced by our CvM by measuring the forward-pass FLOPs on the NYUv2 dataset with an input resolution of $448 \times 576$. Specifically, we evaluate the FLOPs for the SAK-based \cite{lu2024swiss} and DINOv3-based \cite{simeoni2025dinov3} MTL encoder with ViT-L backbone and for our CvM, under the same experimental setting used in the main paper for ViT-L backbone with two input views and a batch size of 1. The FLOPs for the multi-teacher distillation module in SAK is excluded in the calculation. The results are summarized in \cref{supptab:nyuv2forwardcost}. 

When integrated into MTL encoders, our CvM introduces only 0.27 TFLOPs of additional computation, resulting in an increase of approximately 20\% relative to the encoder's original cost, which is a modest computational overhead.
Despite the increased compute, our CvM yields significant performance gains across all tasks, as demonstrated in both quantitative and qualitative evaluations. This trade-off reflects a favorable balance between efficiency and accuracy: the added cost primarily stems from the multi-view transformer and cost volume construction, which inject valuable geometric priors and 3D consistency into the MTL predictions. 
Furthermore, our CvM is designed as a modular, lightweight extension hat can be appended to any existing MTL encoder without requiring architectural changes. Compared to prior work such as 3DMTL \cite{li2023multi}, which incurs a similar level of computational overhead, our CvM provides a more practical solution with better performance for integrating 3D awareness into dense prediction pipelines.

\begin{table}[t]
	\centering
    \resizebox{0.35\textwidth}{!}
    {
		\begin{tabular}{lccccccccc}

            \toprule
            Module & Params. (M) & FLOPs (T) \\
            \midrule
            CvM & $\sim$5 & 0.27 \\
            SAK$_\text{Encoder}$ & $\sim$350 & 1.42 \\
            DINOv3$_\text{Encoder}$ & 300 & 1.23 \\

			\bottomrule
		\end{tabular}%
			}
		\vspace{-0.25cm}
		\caption{Computational cost analysis for CvM with ViT-L backbone. This Table contains both number of parameters for MTL encoder and CvM, and FLOPs for a single forward on NYUv2 dataset.}
		\label{supptab:nyuv2forwardcost}
\end{table}%



\section{More Visualizations}

\begin{figure*}[t]
    \centering
    \includegraphics[width=0.88\textwidth]{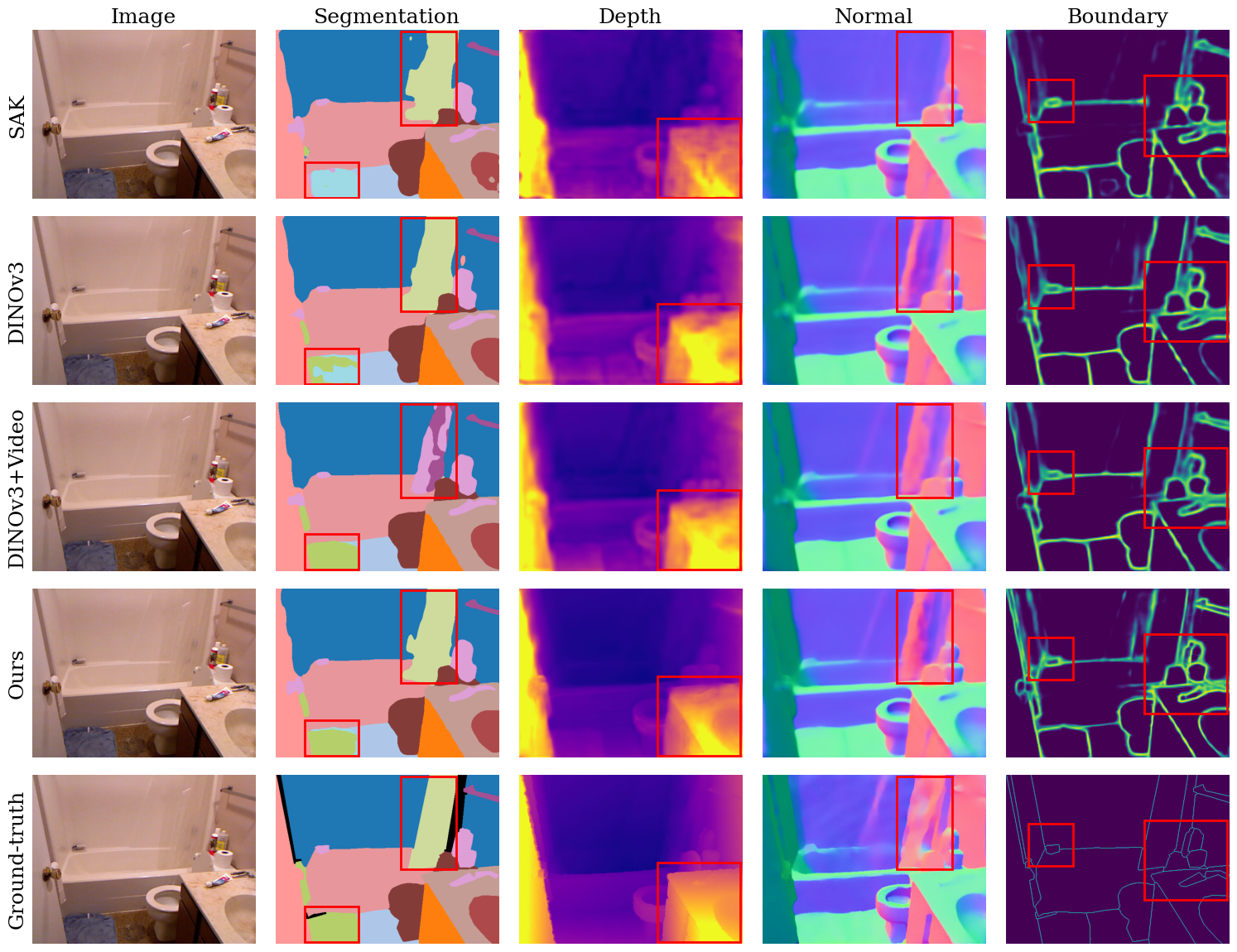}
    \vspace{-0.4cm}
    \caption{ {\bf Qualitative Comparisons on NYUv2.} The first column shows the RGB image, while the remaining columns present either the ground truth or model predictions. The last row shows the ground-truth of four tasks. The first to the fourth row shows the predictions of SAK, Dinov3, Dinov3 trained with videos as multi-view data, and our method, respectively.
    }
    \label{fig:visual_ab}
    \vspace{-0.2cm}
\end{figure*}

\begin{figure*}[t]
    \centering
    \includegraphics[width=0.9\textwidth]{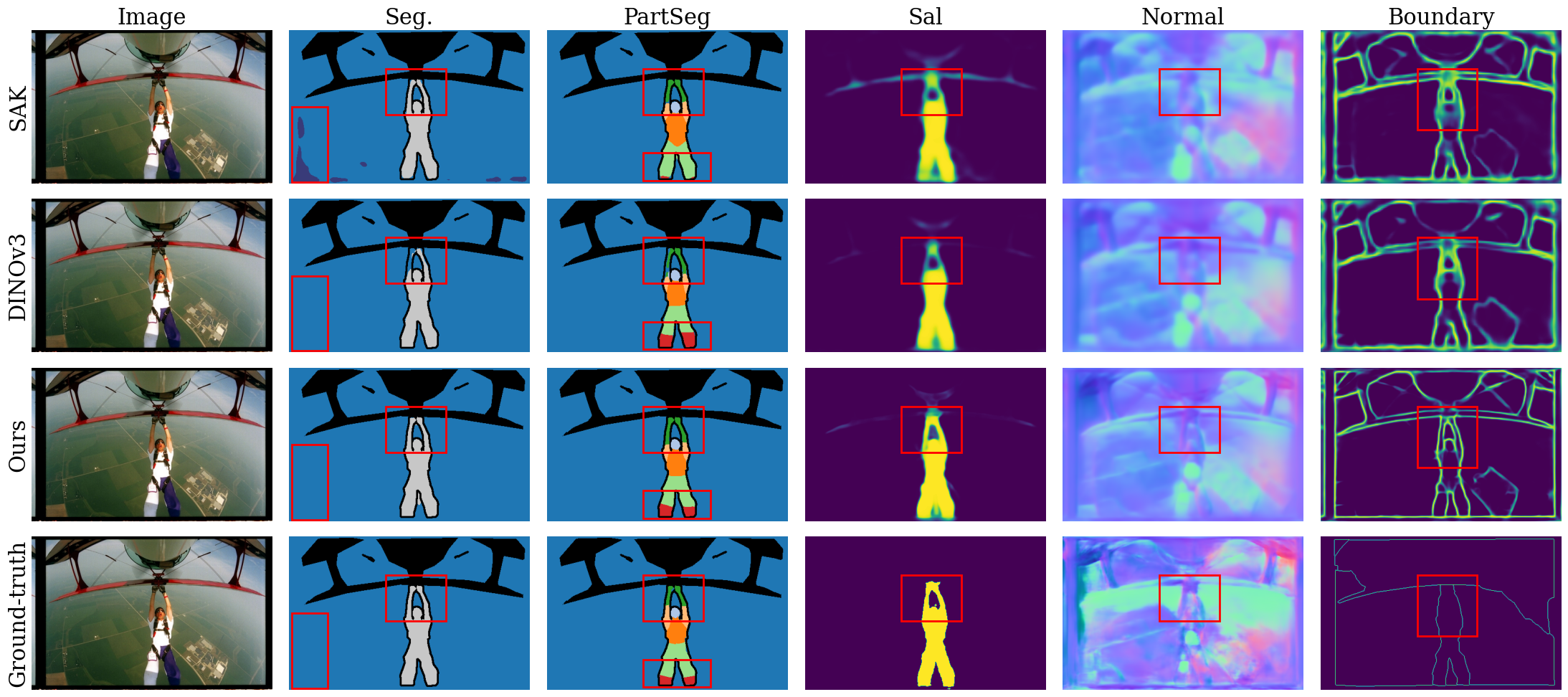}
    \vspace{-0.3cm}
    \caption{{\bf Qualitative Comparisons on PASCAL-Context.} The first column shows the RGB image, while the remaining columns present either the ground truth or model predictions. The last row shows the ground-truth of five tasks. The first to the third row shows the predictions of SAK, Dinov3, and our method, respectively.}
    \label{fig:visual_pascal}
    \vspace{-0.5cm}
\end{figure*}

\begin{figure*}[t]
    \centering
    \includegraphics[width=0.8\textwidth]{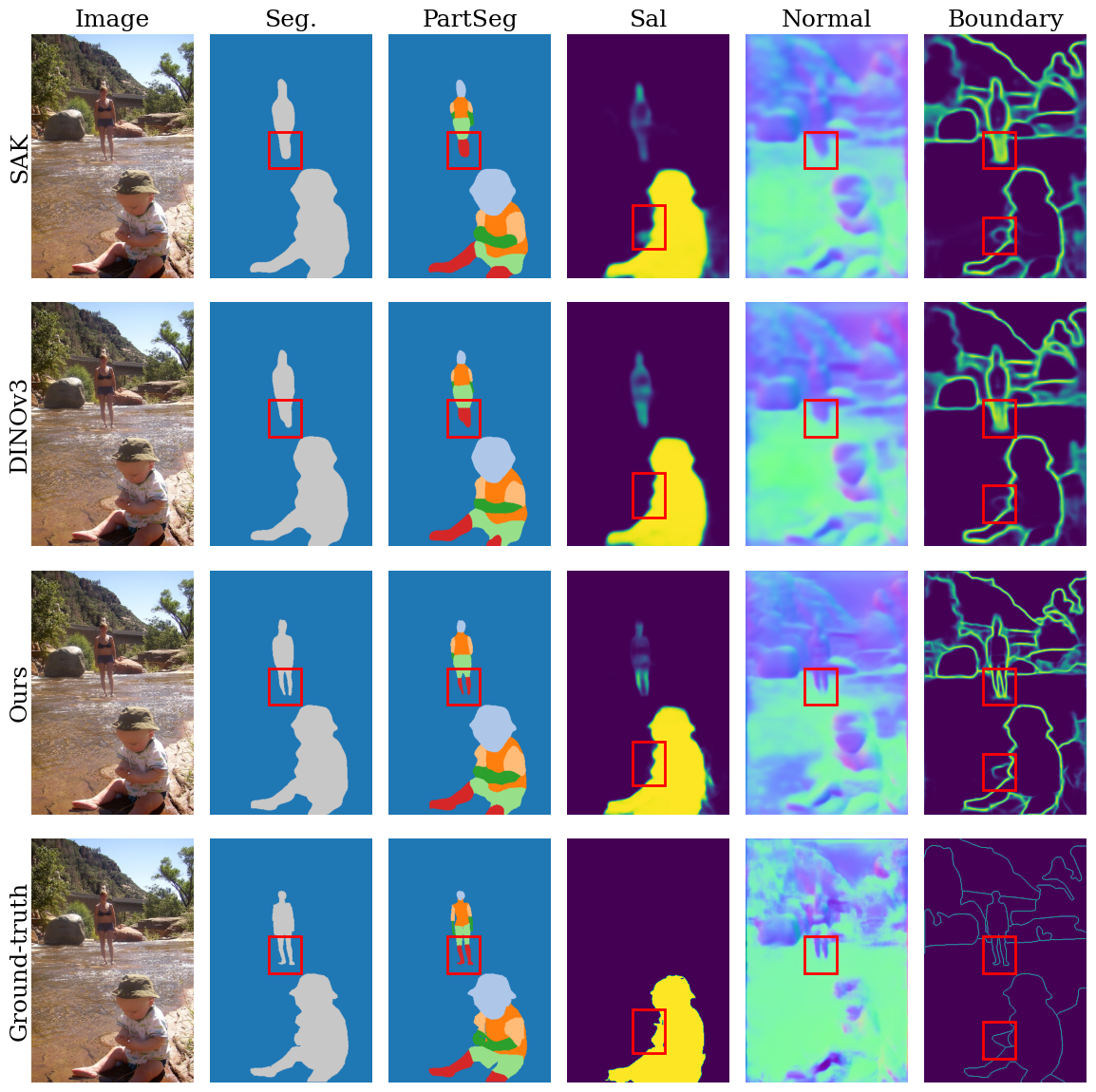}
    \vspace{-0.3cm}
    \caption{{\bf Qualitative Comparisons on PASCAL-Context.} The first column shows the RGB image, while the remaining columns present either the ground truth or model predictions. The last row shows the ground-truth of five tasks. The first to the third row shows the predictions of SAK, Dinov3, and our method, respectively.}
    \vspace{-0.3cm}
    \label{fig:visual_pascal_additional}
\end{figure*}

We provide additional qualitative comparisons between different methods and our method on the NYUv2 and PASCAL-Context datasets. \cref{fig:visual_ab} presents a visualization sample from NYUv2, while \cref{fig:visual_pascal} and \cref{fig:visual_pascal_additional} show two examples from the PASCAL-Context dataset. We visualize ground-truth and predictions of all tasks for each compared methods in the figures. Since PASCAL-Context does not include multi-view video data, we adopt the single-view training setting for these experiments.

Despite the absence of video data for supervising the CvM module, our method still demonstrates clear advantages in semantic tasks, and consistently produces higher-quality predictions for geometric tasks. As shown in \cref{fig:visual_pascal}, for semantic segmentation and human part segmentation, SAK \cite{lu2024swiss} and DINOv3 \cite{simeoni2025dinov3} struggle to distinguish the background from fine-grained regions such as the subject’s arms and legs, while our model successfully recovers these areas. In the saliency task, the traditional MTL model almost collapses the arm into a thin strip, whereas our method preserves the structural integrity of the limb. For edge and surface normal predictions, our CvM also achieves more accurate results, producing high-quality outputs with sharper boundaries and reduced ambiguity around the human body and the control bar of the glider. These results further validate the effectiveness of our CvM and highlight its generalization in single-view settings.


\end{document}